\documentclass[10pt,twocolumn,letterpaper]{article}

\usepackage{cvpr}
\usepackage{times}
\usepackage{epsfig}
\usepackage{graphicx}
\usepackage{amsmath}
\usepackage{amssymb}

\usepackage[ruled,vlined]{algorithm2e}
\usepackage{subcaption}
\usepackage{multirow}
\usepackage{hhline}
\usepackage{array}
\usepackage{bm}

\usepackage{caption}

\newcolumntype{M}[1]{>{\centering\arraybackslash}m{#1}}

\usepackage[breaklinks=true,bookmarks=false]{hyperref}

\cvprfinalcopy 

\def\cvprPaperID{****} 

\ifcvprfinal\fi
\begin{document}

\title{Learning and Refining of Privileged Information-based RNNs for Action Recognition from Depth Sequences}

\author{Zhiyuan Shi, Tae-Kyun Kim\\
Department of Electrical and Electronic Engineering,\\
Imperial College London\\
{\tt\small \{z.shi,tk.kim\}@imperial.ac.uk}
}

\maketitle

\begin{abstract}
Existing RNN-based approaches for action recognition from depth sequences require either skeleton joints or hand-crafted depth features as inputs. An end-to-end manner, mapping from raw depth maps to action classes, is non-trivial to design due to the fact that: 1) single channel map lacks texture thus weakens the discriminative power; 2) relatively small set of depth training data. To address these challenges, we propose to learn an RNN driven by privileged information (PI) in three-steps: An encoder is pre-trained to learn a joint embedding of depth appearance and PI (\ie skeleton joints). The learned embedding layers are then tuned in the learning step, aiming to optimize the network by exploiting PI in a form of multi-task loss. However, exploiting PI as a secondary task provides little help to improve the performance of a primary task (\ie classification) due to the gap between them. Finally, a bridging matrix is defined to connect two tasks by discovering latent PI in the refining step. Our PI-based classification loss maintains a consistency between latent PI and predicted distribution. The latent PI and network are iteratively estimated and updated in an expectation-maximization procedure. The proposed learning process provides greater discriminative power to model subtle depth difference, while helping avoid overfitting the scarcer training data. Our experiments show significant performance gains over state-of-the-art methods on three public benchmark datasets and our newly collected Blanket dataset. 
\end{abstract}
\vspace{-0.3cm}
\section{Introduction}
Action recognition from depth sequences \cite{Yang_tpami_2016,Omar_cvpr_2013,Jiajia_iccv_2013,Vieira_2012,wang2015cnn} has attracted significant interest recently due to the emergence of low-cost depth sensors. Human action refers to a temporal sequence of primitive movements carried out by a person \cite{action_1992}. Recurrent neural network (RNN) \cite{Graves_2012_book} is naturally suited for modeling temporal dynamics of human actions as it can be used to model joint probability distribution over sequences, especially in the case of long short-term memory (LSTM) \cite{Hochreiter_1997} which is capable of modeling long-term contextual information of complex sequential data. 

RNN-based approaches become the dominant solution \cite{wentao_2016,Veeriah_2015_ICCV,Yong_cvpr_2015,Liu2016} for action recognition from depth sequence recently. However, these approaches require either skeleton joints \cite{wentao_2016,Yong_cvpr_2015,Piotr_2016} or hand-crafted depth features \cite{Veeriah_2015_ICCV} as inputs in both training and testing. Skeleton-based action recognition assumes that a robust tracker can estimate body joints accurately in the testing stage. This often does not hold in practice, especially when a human body is partly in view or the person is not in an upright position. Hand-crafted features with heuristic parameters are designed for task-specific data. This often requires multi-stage processing phases, each of which needs to be carefully designed and tuned.

\begin{figure*}[t]
\centering
   \includegraphics[width=\linewidth] {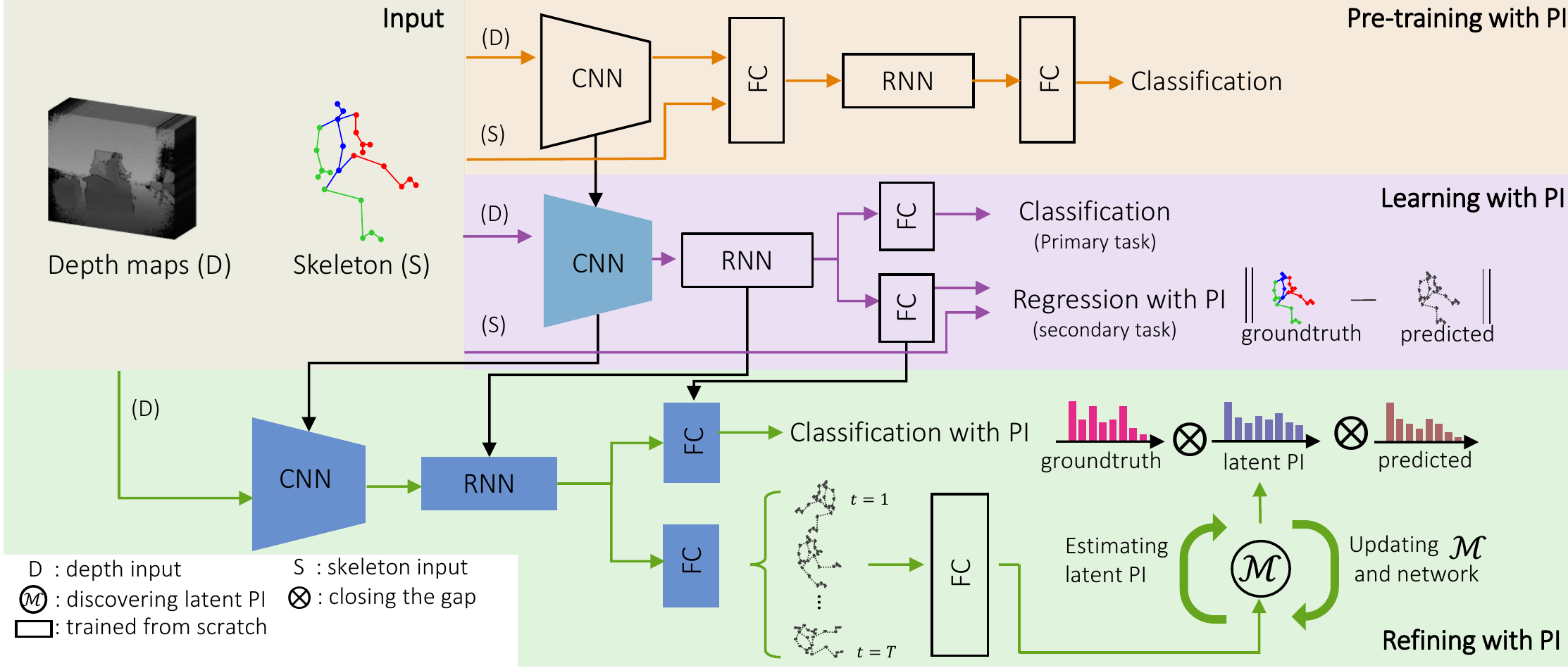}
 
\caption{The proposed framework of PI-based RNNs. Our approach consists of three steps: 1) The pre-training step taking both depth maps and skeleton as input. An embedded encoder is trained in a standard CNN-RNN pipeline. 2) The trained encoder is used to initialize the learning step. A multi-task loss is applied to exploit the PI in the regression term as a secondary task. 3) Finally, refining step aims to discover the latent PI by defining a bridging matrix, in order to maximize the effectiveness of the PI. The latent PI is utilized to close the gap between different information. The latent PI, bridging matrix and the network are optimized iteratively in an EM procedure.}
\label{fig:our_model}
\end{figure*}

An end-to-end trainable model from raw video frames \cite{lrcn2014} is desired to extract spatio-temporal features and model complex sequences in a unified framework. This learning pipeline typically combines a deep convolutional neural network (CNN) \cite{Alex_nips_2012} as visual feature extractor and an RNN \cite{Graves_2012_book} to model and recognize temporal dynamics of sequential data. Unfortunately, these conventional end-to-end manners (CNN+RNN) are difficult to be applied to action recognition from depth sequences due to the fact that: 1) Color and texture are precluded in depth maps, which weaken the discriminative power of the representation captured by the CNN model. 2) Existing depth data of human actions are considered as a small-scale dataset compared to publicly available RGB image dataset. These conventional pipelines are purely data-driven that learn its representation directly from the pixels. Such model is likely at the risk of overfitting when the network is optimized on limited training data.

To address the above-mentioned issues, we propose a privileged information-based recurrent neural network (PRNN) that exploits additional knowledge to obtain a better estimate of network parameters. This additional knowledge, also referred to as privileged information (PI) \cite{Vapnik2009544}, hidden information \cite{Wang_2015_CVPR} or side information \cite{Xu_nips_2013, Hoffman_cvpr_2016}, is only available during training but not available during testing. Our model aims to encode PI into the structure or parameters of networks automatically and effectively during the training stage. In this work, we consider skeleton joints as the PI in the proposed three-step training process (see Fig.~\ref{fig:our_model}). A pre-training stage is introduced that taking both depth sequences and skeleton joints as input. The learned embedding layers construct intermediate distributions over the appearance of depth sequences and skeleton joints. As our method aims to utilize only depth sequences as input in testing stage, we then optimize our model by formulating the PI into an multi-task loss in learning step: a standard softmax classification loss as our primary task, and a regression loss as our secondary task, which learn the mapping parameters to predict the skeleton joints from depth appearance. However, We observe empirically that exploiting PI as a secondary task provides little help to improve the performance of primary task due to the gap between them. Finally, a bridging matrix is defined to connect two tasks by discovering latent PI in the refining step. We present a PI-based classification loss serving as a connector to maintain a consistency between latent PI and primary output distribution by penalizing the violation of the loss inequality. We enforces dependencies across regression and classification targets by seeking shared information. The bridging matrix, latent PI and network parameters are iteratively estimated and updated in an expectation-maximization (EM) procedure.
This proposed learning process can provide greater discriminative power to model subtle depth difference, while helping avoid overfitting the scarcer training data. As we encode skeleton joints as PI, our model does not require a skeleton tracker in a testing stage, showing its better generalizability in a more challenging scenario, such as when a human body is partly in view or the person is not in an upright position. 


We evaluate the proposed PRNN against state-of-the-arts on the task of action recognition from depth sequences. We demonstrate that our approach can achieve higher accuracy on the three public benchmark datasets: MSR Action3D \cite{Wanqing_cvprw_2010}, SBU Interaction dataset \cite{kiwon_cvprw_2012} and Cornell Activity \cite{sung_rgbdactivity_2012}. A larger performance gain can be obtained on our newly collected Blanket dataset, where actions captured from a challenging camera view-point and some actions are partially occluded by a blanket. We also compare with several variants of our model and show that each component consistently contributes to the overall performance.  

 

\section{Related Work}

\noindent \textbf{Action recognition from depth sequence} Human action recognition using depth maps can be classified in local or global methods. The elaborately designed features \cite{Wanqing_cvprw_2010,Jiang_tpami_2014,Omar_cvpr_2013} are typically extracted from spatio-temporal interest points to describe the local appearance in 3D volumes or the area around human joints \cite{Gowayyed_IJCAI_2013}. On the other hand, high-level representations \cite{Xiaodong_2014_cvpr} aim to globally model the postures and capture the temporal evolution of actions. To model sequential state transitions in a principled way, hidden Markov model (HMM) has attracted a lot of interest \cite{gong_tpami_2014} in capturing the temporal structure of human action dynamics. These HMM-based methods require that video sequences are precisely cropped and aligned with actions of interest, which itself is a difficult task for real-world videos. RNNs are able to handle both variable-length input and output that become the dominant model \cite{Veeriah_2015_ICCV,Yong_cvpr_2015,wentao_2016} recently, achieving superior performance over previous approaches. HBRNN \cite{Yong_cvpr_2015} divides human skeleton into five corresponding parts and feed them into five bidirectionally recurrently connected subnets. \cite{wentao_2016} improve the model of \cite{Yong_cvpr_2015} by automatically discovering the inherent correlations among skeleton joints. Instead of assuming skeleton joints are always reliable in testing stages, \cite{Veeriah_2015_ICCV} model the dynamic evolution of actions by measuring the salient motions from the input depth appearance. The depth features are still extracted based on hand-crafted heuristics. In this paper, we provide an end-to-end solution to action recognition from raw depth sequences.  
 
\vspace{0.2cm}
\noindent \textbf{Learning with PI} Data-driven approaches leverage large amounts of training data to determine the optimal model parameters in a bottom-up fashion. Purely data-driven methods are often very brittle and prone to fail when learning with limited training data, due to overfitting or an optimization obstacle involved. Learning with additional knowledge is a natural solution to alleviate this issue. This knowledge, also referred to as PI \cite{Vapnik2009544}, hidden information \cite{Wang_2015_CVPR} or side information \cite{Xu_nips_2013}, which can help to provide more explanations in training but will not be available at testing. Learning with PI has been investigated in many existing algorithms. \cite{Feyereisl_2014_nips} incorporate PI into an objective function of a structural SVM to improve object localization performance. \cite{dantone_2012_cvpr} show that the incorporation of additional information can enhance the dependency between output variables and latent variables in a random forest framework. Additional knowledge has also been considered in neural networks. \cite{JMLR_gulchere16a} explore the architecture by providing intermediate targets. \cite{eslami_air_2016} demonstrate the effectiveness of prior distribution for adjusting the model parameters to improve its generalization. More recently, \cite{Mahasseni_2016_CVPR} present a regularized RNNs with additional information for RGB video sequences. However, PI is either pre-trained or fixed in previous methods. In this work, we propose to optimize our end-to-end trainable model with iteratively estimating and updating latent PI for depth-based action recognition.

\section{Spatio-Temporal Modeling}

We illustrate an overall view of our model in Figure \ref{fig:our_model}. The architecture mainly consists of an encoder, recurrent layers and PI-based learning. The encoder consists of several layers of convolutions which takes as input a collection of videos $\mathcal{V}$, where each video $V_j$ is a sequence of frames  $V_j=\{v_t : t=1,...,T_j\}$. The encoder produces vector space representations $X_j=\{\bm{x}_t : t=1,...,T_j\}$ for all frames of $V_j$. The recurrent network is built for integrating over time all the available information from $X_j$. Finally, PI is incorporated to jointly optimize all the layer parameters in the proposed three-step learning process.

\noindent \textbf{Convolutional Neural Network} The spatial appearance of action and contextual scenes on an individual frame is captured by our encoder. The architecture of our encoder is illustrated in Figure \ref{fig:encoder}. It is inspired from VGG-VeryDeep \cite{Simonyan15c}, which is slightly modified from the 11 weights layer version by considering the depth maps and smaller training data. The network comprises five convolutional layers, five max-pooling layers. The rectified linear unit \cite{Alex_nips_2012} is adopted as the activation function. Compared to the widely used CNN encoder for RGB data \cite{Mahasseni_2016_CVPR,Simonyan15c}, our encoder is more compact and effective for depth sequences. It is used to extract a feature vector from an input frame. Given an input depth frame $\bm{v}_t \in \mathbb{R}^{224\times224}$, an activation map $f_t^6 \in \mathbb{R}^{7\times7\times512}$ can be obtained from ``outMap6'' layer. We apply a linear transformation between the activation map and feature vectors by $\bm{x}_t=\mathrm{tanh}(\bm{W}^6f_t^6+b^6)$. This ``map to sequences'' operation generates an input vector $\bm{x}_t \in \mathbb{R}^{1\times1000}$ for recurrent layers in refining step.

\noindent \textbf{Recurrent Neural Network} RNNs are neural networks with feedback loops that produce the recurrent connection in the unfolded network \cite{Chung_NIPS2015,Jonas_2016_aaai,Qiang_aaai_2016}.
Given an input sequence from the above encoder $X_n$, the hidden states of a recurrent layer $h_j = (\bm{h}_t : t=1,...,T_j)$ are defined as  $\bm{h}_{t}=\mathrm{tanh}(\bm{W}^{h}\bm{x}_{t}+\bm{U}^{h}\bm{h}_{t-1}+b^h)$. Here $\bm{W}^h$,$\bm{U}^h$ are parameters of an affine transformation which update the connection weights among input layer, hidden layer. RNNs suffer from the vanishing and the exploding gradient problem \cite{Bengio_1994}. We adopt LSTM \cite{Hochreiter_1997} to address the problem of learning long-range dependencies, where a memory cell vector $\bm{c}_t$ is maintained at each time step $t$. 
LSTM contains one self-connected memory cell $\bm{c}$ and three multiplicative units, \ie the input gate $\bm{i}$, the forget gate $\bm{f}$ and the output gate $\bm{o}$, which can store and access the long range contextual information of a temporal sequence. Please refer to \cite{Hochreiter_1997} for the precise form of the update. 

\section{PI-Based RNNs}

Standard recurrent neural networks do not provide a mechanism to exploit the PI when it is available at training time. We first present a pre-training strategy. The learned encoder is applied to the learning step and  tuned together with RNNs by formulating the PI into a multi-task loss. In the final refining step, latent PI is discovered and iteratively updated with network parameters.


\subsection{Pre-training with PI}
\label{sec:pretrain}

A pre-training strategy is proposed to learn a joint embedding by taking both depth sequences $V_j$ and skeleton joints annotation $\bm{E}=\{ \bm{e}_1,...,\bm{e}_S \}$ as input. Each $\bm{e}_s \in \mathcal{R}^3$ has 3 coordinates. In this stage, $\bm{x}_t$ is not directly applied to RNNs. Instead, the additional layer transforms $\bm{x}_t$ together with $\bm{E}$ to derive an embedding space : 
\begin{align}
\bm{x}^{\prime}_t=\mathrm{tanh}(\bm{W}^7\bm{x}_t+\bm{W}_{e}\bm{E}+b^7)
\label{eq:pretrain}
\end{align}
where $\bm{W}_{e}$ is the weight matrix connecting the skeleton joints. The resulting $\bm{x}^{\prime}_t$ have the same dimensionality (1000) as $\bm{x}_t$. This is followed by RNNs to model the dynamics of sequential data. Finally, similar to most RNNs for classification task, a softmax layer is adopted to transform the hidden state vector into the probability distribution of action classes.

The key insight of the pre-training stage is to learn a depth encoder that optimizes the embedding over both depth appearance and skeleton joints. The learned encoder serves as an initialization in the next learning stage. This pre-training stage leads to a significant improvement in both efficiency and effectiveness.

\subsection{Learning with PI}
\label{sec:learning}

\noindent \textbf{Multi-task loss.} To obtain the class predictions of an input sequence $\bm{X}_j$, the hidden state can be mapped to an output vector $\bm{y}_j=(\bm{y}_t: t=1,...,T_j)$. During training, we measure the deviation between groundtruth and last memory cell at the frame $T$ for classification loss, since LSTMs have the ability to memorize the content of an entire sequence. For regression loss, we accumulate the loss of each frame t across the $T$ frame sequence. The final objective function in the learning step is to minimize the cumulative maximum-likelihood loss over all training sequences: 
\vspace{-0.05cm}
\begin{align}
\mathcal{L}^{\mathrm{L}}(\bm{\Omega})=& 
\sum_{j=1}^{J}\mathcal{L}^{c}(T,j)+ \lambda \sum_{j=1}^{J}\sum_{t=1}^{T}\mathcal{L}^{r}(t,j)\label{eq:joint_loss}
\end{align}
There are $J$ sequences in the training set $\bm{\Omega}$. 
The hyper-parameter $\lambda$ in Eqn.~\ref{eq:joint_loss} controls the balance between the two losses. The classification loss and regression loss are defined as follows:

\noindent \textbf{Classification loss.}  $\bm{y}_t \in \mathbb{R}^K$ represents an 1-of-K encoding of the confidence scores on $K$ classes of actions, which can be derived as  $\bm{y}_{t}=\mathrm{tanh}(\bm{W}^{y}\bm{h}_{t}+b^{y})$. This output vector can be transformed into a vector of probabilities $p(y_{tk})$ for each class $k$ by softmax function as $p(y_{tk})=e^{y_{tk}}/\sum_{l=1}^{K}e^{y_{tl}}$. To learn the model parameters of our model, cross entropy loss between the predicted distribution $p(\bm{y}_{t})$ and target class $g_t$ is defined as 
$$\mathcal{L}^{c}(t,j)( =- \sum_{k=1}^{K} \delta (k-g_t)\log p(y_{jtk})$$ for the sample $t$ of the $j$-th video,
where $\delta(\cdot)$ is the Dirac delta function, and $g_t$ denotes the groundtruth label of the sample $t$.

\noindent \textbf{Regression loss.} Besides classification output, our model has another sibling output layer as regression term. We define a skeleton regression targets for groundtruth keypoints $\hat{\bm{E}}_t=\{ \hat{\bm{e}}_{t1},...,\hat{\bm{e}}_{tS} \}$  and predicted locations $\bm{B}_t=\{\bm{b}_{t1},...,\bm{b}_{tS}\}$ at each time step $t$. We select $\hat{\bm{E}}$ as a subset of the skeleton annotations $\bm{E}$,  because this is secondary target and an accurate estimation of all skeleton joints is not needed in testing. Each instance is accompanied with a set of keypoint $\{\hat{e}^x_{ts}, \hat{e}^y_{ts}\}_{s=1}^S$ locations, which are normalized with respect to the center and the width and height of the input region. The loss associated with the task of measuring the skeleton estimation can be expressed as 
$$\mathcal{L}^{r}(t,j)=\frac{1}{S}\sum_{s=1}^{s=S}((\hat{e}^x_{jts}-b^x_{jts})^2+(\hat{e}^y_{jts}-b^y_{jts})^2)$$
where we use $L_2$ distance between the normalized keypoints location to quantify the dissimilarity. This loss function and regression layer only appear in the training stage for optimizing the neural network with additional information.

This extension, known as multi-task learning \cite{Mitchell_1997}, utilize the task relationships to learn all individual tasks simultaneously, such that information can be shared in the common structure of the model to benefit all tasks. Similar as \cite{girshick15fastrcnn}, it will help the classification prediction by considering the regression aspects. During testing, the regression component will be disabled.

\subsection{Refining with PI}
\label{sec:refine}

However, the conventional multi-task loss in the last step does not consider any relationship between two tasks. We observe empirically that purely exploiting PI as a secondary task provides little help to improve the performance of primary task due to the gap between them. To maximize the effectiveness of PI for helping primary task, we propose to discover latent PI from the secondary task in this refining step. The latent PI is utilized in the primary task to optimize the network. The updated network is further used to refine latent PI iteratively in an EM procedure. 

\noindent \textbf{Latent PI modeling}  We define latent PI as a informative distribution which is jointly modeled by secondary task and a bridging matrix. The bridging matrix $\bm{\mathcal{M}}$ aim to capture the underlying dependencies between primary and secondary task. The log-likelihood of the defined model can be expressed as:
\begin{align}
Q(\bm{\Theta},\bm{\mathcal{M}}) = \sum_{j=1}^{J}\log(\sum_{k=1}^{K}p(y^{\prime}|\bm{X}_j;\bm{\Theta})p(g_j|y^{\prime};\bm{\mathcal{M}})),
\end{align}

where $\bm{\Theta}$ is the set of parameters of the network in refining step. Given $\bm{\Theta}$, which initialized by the model from the learning step, we can predict the skeleton joints $\bm{B}_t$ of a depth frame. We concatenate the predicted skeleton of every frame to a single vector $\bm{B}=\{\bm{B}_{1},...,\bm{B}_{T_n}\}$. $y^{\prime}$ is then calculated as a fully connected layer: $\bm{y^{\prime}}=\bm{W}^{y^{\prime}}\bm{B}+b^{y^{\prime}}$. $\bm{W}^{y^{\prime}}$ and $b^{y^{\prime}}$ is part of $\bm{\Theta}$, but they are trained from scratch. During the training of the refining step, our model aims to maximize the likelihood function by optimizing both the bridging matrix and network parameter iteratively in an EM procedure.

\noindent \textbf{Estimating latent PI} The explicit expression of latent PI is as follows: 
\begin{align}
u_k &= p(y_k^{\prime} | \bm{B}, g ; \bm{W}_{y^{\prime}}, \bm{\mathcal{M}}) \nonumber \\
&= \frac{p(g|y_k^{\prime};\bm{\mathcal{M}})p(y_k^{\prime}|\bm{B}; \bm{W}_{y^{\prime}})}{\sum_{l=1}^{K}p(g|y_l^{\prime};\bm{\mathcal{M}})p(y_k^{\prime}|\bm{B};\bm{W}_{y^{\prime}})} \nonumber \\
&= \frac{\bm{\mathcal{M}}_{kg}\exp(\bm{W}_k^{y^{\prime}}\bm{B}+b_{y^{\prime}})}{\sum_{l=1}^{K}\bm{\mathcal{M}}_{lg}\exp(\bm{W}_k^{y^{\prime}}\bm{B}+b_{y^{\prime}})}
\label{eq:estep}
\end{align}

$p(y_k^{\prime}|\bm{B}; \bm{W}_{y^{\prime}})$ is a predicted probability of the class k by observing the predicted skeleton joints $\bm{B}$ of a input depth sequences. The bridging matrix $\bm{\mathcal{M}}$ aims to transform the predicted distribution to a latent distribution that can be effectively used in optimizing the network.

\setlength{\textfloatsep}{0.3cm}

\begin{algorithm}[t]
\SetKwBlock{pretrain}{Pre-training:}{end}
\SetKwBlock{learning}{Learning:}{end}
\SetKwBlock{refining}{Refining:}{end}
\SetKwBlock{estep}{E-step:}{end}
\SetKwBlock{mstep}{M-step:}{end}
\SetKwBlock{Beg}{ddd}{rrr}
\SetAlgoLined
\KwIn{A collection of videos $\mathcal{V}$, skeleton joints annotation $\bm{E}$, subset of skeleton joints $\bm{\hat{E}}$, groundtruth class label $g$.} 

\KwOut{Network parameters, bridging matrix $\bm{\mathcal{M}}$}

\pretrain{
Eq.\ref{eq:pretrain} taking both $\bm{x}$ of depth sequences $V$ and skeleton joints $\bm{E}$, \\
A encoder is trained by minimize the standard softmax loss.
}	

\learning{
Taking the subset of skeleton joints $\bm{\hat{E}}$ in the regression term. \\
The parameters of network are optimized by minimizing the multi-task loss Eq.~\ref{eq:joint_loss}
}

\refining{
\While{not converge}{
\estep{
Estimating and updating the latent PI by Eq.~\ref{eq:estep}
}
\mstep{
The parameters of network are optimized by PI-based classification loss Eq.~\ref{eg:refine_loss}. \\
The bridging matrix $\bm{\mathcal{M}}$ is updated with Eq.~\ref{eg:bridge_matrix}

}
}
}
\caption{\label{alg:prnn}PI-based RNNs}
\end{algorithm}

\setlength{\belowcaptionskip}{-8pt}

\noindent \textbf{Updating model with latent PI} The distribution of latent PI $p(\hat{\bm{u}}_j)$ of an input sequence $\bm{X}_j$ is defined by $p(\hat{\bm{u}}_j)=\bm{u}_j \bm{z}_t$, where $\bm{z}_t \in \mathbb{R}^K$ is randomly generated for each frame $t$ from a Multinoulli distribution $ \{ \hat{g} \sim \mathcal{P}(\alpha),  z_{\hat{g}} =1, z_l=0,  \forall l \neq g \}$, where $\mathcal{P}(\alpha)$ is defined as $p_g = 1- \frac{K-1}{K}\alpha$ and $p_l = \frac{1}{K}\alpha$, where $\alpha$ is to control how strongly the prior distribution is pushed to classification loss, and $g$ is the groundtruth label. 
We replace the groundtruth label by the probabilities of latent PI to formulate the PI-based classification loss in refining step: 
\vspace{-0.05cm}
\begin{align}
\mathcal{L}^{\mathrm{R}}=-\sum_{j=1}^{J}\bigg(\sum_{k=1}^{K} p(\hat{u}_{jk})\log p(y_{jTk}) \nonumber \\
-\beta\sum_{k=1}^{K} \delta(k-g_j) \log p(y_{jk}^{\prime})\bigg)
\label{eg:refine_loss}
\end{align}

a standard softmax loss is also included in $\mathcal{L}^{\mathrm{R}}$ to update the parameters (\eg $\bm{W}^{y^{\prime}}$, $b^{y^{\prime}}$) from the branch of secondary task. Apart from optimizing network parameters, the bridging matrix of modeling latent PI can be updated iteratively by a closed-form solution in the M-step of EM procedure \cite{mclachlan2007algorithm, Shi_2015_tpami,Alan_2016_icassp}:

\begin{align}
\bm{\mathcal{M}}_{kl}(\Omega) = \frac{\sum_{j=1}^{J} u_{jk} \delta(l-g_j)}{\sum u_{jk}}, \ \ \qquad  k,l \in \{1,...,K\}
\label{eg:bridge_matrix}
\end{align}

\begin{figure*}[t]
\centering
   \includegraphics[width=\linewidth] {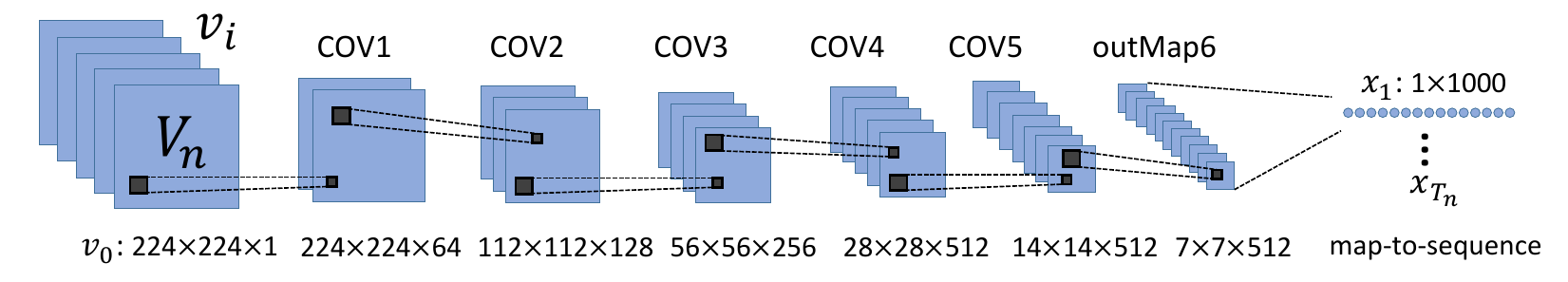}
 
\caption{The architecture of the encoder. The convolutional layers (from COV1 to COV5) with kernel size 3$\times$3 and a stride of 1. The padding implements “same” convolution (and pooling), where the input and output maps have the same spatial extent. max-pooling is performed from COV1 to COV5 over 2$\times$2 spatial windows with stride 2.}
\label{fig:encoder}
\end{figure*}

\noindent \textbf{Discussion on latent PI} Latent PI can be treated as a sufficient information to act as a teacher network \cite{Korattikara_nips_2015,Szegedy_2015}. However, our latent PI is obtained in the same framework rather than trained from a separate model. Our model further refines latent PI according to the feedback of the network in each iteration. 
This updating process us two benefits: (1) The formulation strikes a good balance between the class distributions learned from depth appearance and skeleton information.  This is similar in spirit to \cite{Schulter_cvpr_2013}, where a weight distribution is utilized to improve the learning process of random forest. Sun \etal \cite{Sun_cvpr_2012} also incorporate prior information (\eg human height) to enhance the dependency between output variables and latent variables, where the prior can help to split data effectively. The skeleton and raw depth sequence should share relevant and complementary information.  Here, we measure the loss by partially considering the posterior obtained from skeleton joints. We show that this learning process improves the discriminative power of the network. (2) Apart from learning better depth representation, our PI-based classification loss provides an effective way to prevent overfitting. Since the prior label is not perfectly trained, the noise is introduced when we switch to the prior label according to $\alpha$. This term can be treated as a regularizer similar as \cite{xie2016disturblabel}, where they intentionally generate incorrect training labels at the loss layer. Our loss function also seeks to minimize the confusion between the two distributions.



\subsection{Model Training}

We summarize the whole training process of the proposed PI-based RNNs in Algorithm~\ref{alg:prnn}. Note that the learning step and refining step can be potentially preformed alternatively to improve the effectiveness of the trained model. In our experiments, we show that one round of learning and refining step achieves significant improvements. While small improvements can be further obtained with more rounds, which has been verified on SBU dataset, we fix to one round of learning and refining for all experiments with the good trade-off between accuracy and efficiency. In refining step, the EM procedure is still run iteratively until convergence. 

For all three steps, the error differentials measured by the last layer of the recurrent neural network will be back-propagated to feature sequences and feed back to the convolutional layers across every frame in the videos. Our approach is an end-to-end trainable network that jointly learns the parameters of the CNN and the RNN. We train each model with stochastic gradient descent on the negative log-likelihood using the Adam optimizer, with a learning rate of 0.001 for MSR Action3D and 0.0001 for the rest. A minibatch size of 10 is applied to all datasets. We use early stopping when the validation error starts to increase.

\section{Experiments}
We compare the performance of our model with state-of-the-art methods and baselines on four datasets: MSR Action3D Dataset \cite{Wanqing_cvprw_2010} (Action3D), SBU Interaction dataset \cite{kiwon_cvprw_2012} (SBU), Cornell Activity Dataset \cite{sung_rgbdactivity_2012}(CAD60), and the proposed Blanket dataset (Blanket). We also analyze each component of our model and the computational efficiency.

\noindent \textbf{Datasets:} \textbf{Action3D} is an action dataset of depth sequences captured by a depth camera.  This dataset consists of 20 actions performed by 10 subjects. Every action was performed by ten subjects three times each. All sequences are captured in 15 FPS, and each frame in a sequence contains 20 skeleton joints. Altogether, the dataset has 557 valid action sequences with 23797 frames of depth maps. \textbf{SBU} consists of 282 pre-segmented sequences, which includes 8 classes depicting two-person interaction. Each action is performed by 21 pairs of subjects. \textbf{CAD60} consists of 68 video clips captured by Microsoft Kinect device. Each video is of length about 45s. Four different subjects performed 14 different activities in five locations: office, kitchen, bedroom, bathroom and living room. \textbf{Blanket} contains 120 depth video clips. There are 12 different action classes performed by 10 subjects. Our dataset contains more static actions (\eg lying and sitting). This dataset is very challenging, as some actions are partially occluded by a blanket. For example, one actor is sitting on the bed while he is covered by a blanket (please refer to our supplementary video for all actions).

\noindent \textbf{Implementation details:} We implemented the network using TensorFlow \cite{tensorflow}. The architecture of convolutional layers (see Fig.~\ref{fig:encoder}) is slightly modified from VGG-VeryDeep \cite{Simonyan15c} (with 11 weight layers) for depth maps.  We initialize the weights without pre-training by using the normalized initialization procedure \cite{glorot2010understanding}. Unlike images which can be rescaled and randomly cropped to a fixed size, spatio-temporal consistency has to be considered for video sequences. Each input video frame is scaled to 227x227 from the whole frame. We did not perform the operation of randomly cropped and flipped for utilizing PI easily. The depth values are normalized to [-1,1]. Our model has a stack of 2 LSTMs of 1000 hidden units each. To reduce the computation cost, we sample each video of CAD60 with a maximum length of 200 frames. We do not sample frames from MSR Action3D, SBU and Blanket dataset. We unroll the LSTM to a maximum length of 200 time steps for CAD60, 300 time steps for Blanket and 100 time steps for the rest during training, which is a good trade-off between accuracy and complexity. 

We mainly consider the skeleton joints as our PI. The prior class distribution is obtained by training DURNN-L \cite{Yong_cvpr_2015} with all available skeletons. In our regression loss, we use only six joints (\ie head, hand left, hand right, foot left, foot right, hip center) as this secondary target is formulated for helping classification accuracy. For our Blanket dataset, we annotate the six joints for both pre-training and refining stage because of the special camera view-point. We normalize 3D joint coordinates to a unified coordinate system from the world coordinate system by placing the hip center at the origin \cite{Vemulapalli_cvpr_2014}. Similar as \cite{Yong_cvpr_2015},  we apply a simple Savitzky-Golay smoothing filter to smooth the skeleton annotations.  

\begin{figure}[t]
\centering
   \includegraphics[width=\linewidth] {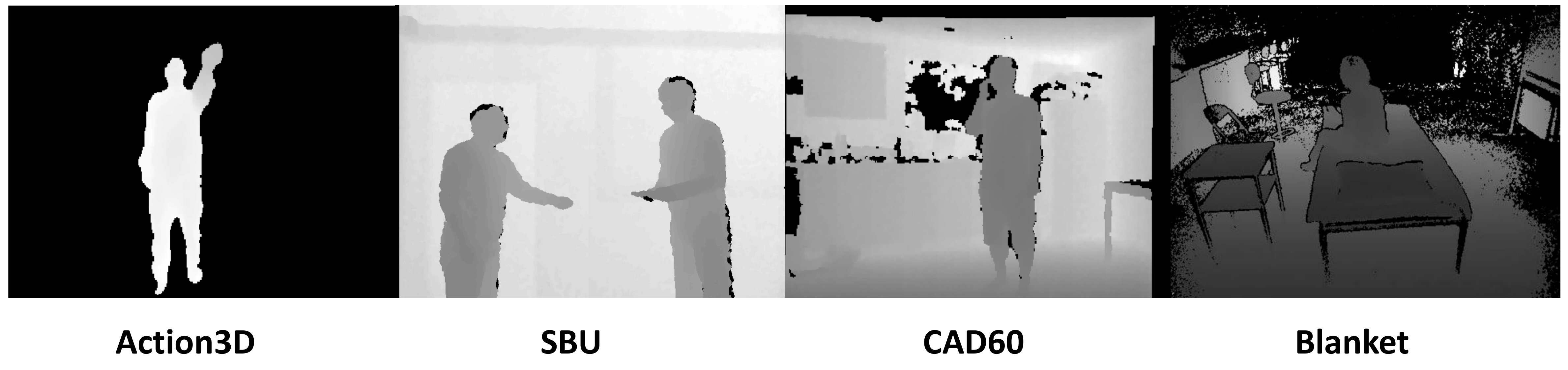}
 
\caption{Examples of depth maps on four datasets.} 
\label{fig:data}
\end{figure}

\vspace{-0.1cm}
\subsection{Comparison to the State-of-the-art}
\vspace{-0.1cm}
The experimental results are shown in Table~\ref{tab:soa}. Existing state-of-the-art methods can be partitioned into two groups: using 1) only the depth sequences or 2) at least skeleton information in the testing stage.

\noindent \textbf{Results on Action3D :} We follow a similar evaluation protocol from \cite{Chunyu_aaai_2016,Wang_2016_CVPR}. In this setting, the dataset is divided into two sets, where half of the subjects are used for training and the other half are used for testing. Compared to another protocol \cite{Yong_cvpr_2015} that splits classes into three subsets, this setting is more challenging as all actions are evaluated together. 
The average accuracy corresponds to the mean of the confusion matrix diagonal of all classes. Note that 10 skeleton sequences were not used \cite{Jiang_tpami_2014} because of missing data. We compare the proposed model PRNN with Xia \etal \cite{Lu_cvpr_2013}, Oreifej \etal \cite{Omar_cvpr_2013}, and Yang \etal \cite{Xiaodong_2014_cvpr}. All theses methods require only depth maps as input during testing. We can see that our proposed PRNN achieves the best average accuracy (94.9 \%) compared with them. For a complete comparison, we also list those skeleton-based approach in the lower part of the Table~\ref{tab:soa}. Skeleton-based approaches demonstrate slightly better performance by assuming a robust skeleton tracker is available in testing. Our method aims to provide a more general framework allowing us to learn the model directly from raw observations of depth videos, rather than explicitly modeling skeletal joints \cite{Yong_cvpr_2015} or local appearance \cite{Veeriah_2015_ICCV}. Many of these methods either focus on modeling spatio-temporal structure with a certain assumption \cite{Omar_cvpr_2013}, or exploit the trajectories of human joints \cite{Veeriah_2015_ICCV,Zanfir_2013_ICCV} in the testing stage which rely on accurate skeleton joints detection. 

\noindent \textbf{SBU :} We follow the experimental setting of \cite{kiwon_cvprw_2012,wentao_2016} and use five-fold cross-validation. All action categories are composed of interactions between actors, involving human acting and reacting. This dataset is very challenging, especially in our setting where skeleton information is not available in testing. We summarize the results in Table~\ref{tab:soa}. We can see that our method achieves superior performance to the depth-based approaches and perform close to the skeleton-based approaches. 

\noindent \textbf{CAD60 :} We follow the same experimental setting as in \cite{Jiang_tpami_2014,CVPR15_heterogeneous} by adopting the leave-one-person-out cross validation. \ie the model was trained on three of the four people, and tested on the fourth. Table~\ref{tab:soa} compares the results on CAD60. 
We can see that the proposed PRNN achieves the 87.6\% accuracy with only seeing the depth maps, comparing against previous works which utilize multiple cues (\ie RGB frames, depth maps and the tracked skeleton joint positions) 
in testing. Some different human actions of CAD60 share similar body motions such as ``chopping'' and ``stirring''. Our model takes advantage of the PI-based learning process, which allows to distinguish the subtle motions from depth maps \cite{Chengcheng_aaai_2014}.

\begin{table}[t]
\setlength{\tabcolsep}{0.3em}
\setlength\extrarowheight{0.3pt}
\centering
\footnotesize
\scalebox{0.95}{
\begin{tabular}{M{0.5cm} l || l  ||  l ||  l || l }
\hline
\multicolumn{2}{c||}{\multirow{1}{*}{Method}} & \multicolumn{1}{c||}{Action3D}  &\multicolumn{1}{c||}{SBU} & \multicolumn{1}{c||}{CAD60} & \multicolumn{1}{c}{Blanket} \\
\hline
\hline

\parbox{5mm}{\multirow{4}{*}[-3pt]{\rotatebox{90}{depth}}} &  \multicolumn{1}{|l||}{Xia \etal \cite{Lu_cvpr_2013}}  & 89.3 & 43.69 & - & 40.6 \\
\cline{2-6}
& \multicolumn{1}{|l||}{Oreifej \etal \cite{Omar_cvpr_2013}}  & 88.9 & 77.0 & 72.7  & 42.8 \\
\cline{2-6}
& \multicolumn{1}{|l||}{Yang \etal \cite{Yang_tpami_2016}}  & 93.45 & - & -& 41.2 \\
\cline{2-6}

& \multicolumn{1}{|l||}{\textbf{PRNN}}  &  \textbf{94.9} & \textbf{89.2} & \textbf{87.6} & \textbf{53.5}  \\
\hhline{======}
\parbox{5mm}{\multirow{10}{*}[-3pt]{\rotatebox{90}{skeleton}}}
& \multicolumn{1}{|l||}{Vemulapalli \etal \cite{Vemulapalli_cvpr_2014}}  & 89.48 & - & - & - \\
\cline{2-6}
& \multicolumn{1}{|l||}{Veeriah \etal \cite{Veeriah_2015_ICCV}}  & 92.03 & - & - & - \\
\cline{2-6}
& \multicolumn{1}{|l||}{Hu \etal \cite{CVPR15_heterogeneous}}  & - & - & \textbf{84.1} & - \\
\cline{2-6}
& \multicolumn{1}{|l||}{Koppula \etal \cite{Koppula_2013}}  & - & - & 71.4 & - \\
\cline{2-6}
& \multicolumn{1}{|l||}{Du \etal \cite{Yong_cvpr_2015}}  & - & 80.35 & - & - \\
\cline{2-6}
& \multicolumn{1}{|l||}{Wang \etal \cite{Wang_2015_ICCV}}  & \textbf{96.9} & - & - & - \\
\cline{2-6}
& \multicolumn{1}{|l||}{Wang \etal \cite{Chunyu_aaai_2016}}  & 91.40 & - & - & - \\
\cline{2-6}
& \multicolumn{1}{|l||}{Zhu \etal \cite{wentao_2016}}  & - & 90.41 & - & - \\
\cline{2-6}
& \multicolumn{1}{|l||}{Gori \etal \cite{Gori_2016}}  & 95.38 & \textbf{93.08} & - & - \\
\cline{2-6}
& \multicolumn{1}{|l||}{Wang \etal \cite{Jiang_tpami_2014} }  & 88.2 &- & 74.7  & - \\


\hline
\hline
\end{tabular}
}

\caption{Comparison with state-of-the-art methods on four datasets for action recognition. '-' indicates no result was reported and no code is available for implementation.}
\label{tab:soa}
\end{table}

\noindent \textbf{Blanket:} Similar to CAD60, we follow the protocol as \cite{Jiang_tpami_2014} and perform cross-validation on our proposed dataset. We compare our model with three baseline methods:  Xia \etal \cite{Lu_cvpr_2013}, Oreifej \etal \cite{Omar_cvpr_2013}, Yang \etal \cite{Yang_tpami_2016}. We use their publicly available codes and train their model with varying their parameters, so as to report the best results for fair comparison. The experimental results are shown in Table~\ref{tab:soa}. The proposed PRNN obtains the state-of-the-art accuracy of 53.5\%. Our collected data is more difficult to learn than the existing dataset. Although each basic action is simple like ``sitting'' and ``lying down'', the actor (\ie patient) is either partially occluded by a blanket or in a suffering status when he performs these actions. It introduces severe noise (\eg shaking his body, trembling) to the basic actions. Moreover, this special camera view-point (see Figure~\ref{fig:data}) and the occlusion by a blanket will cause difficulties for skeletal estimation. As expected, a larger performance gap is seen between our model and other approaches. This demonstrates the potential of our model in representing and modeling the dynamics of actions directly from depth maps. 

In brief, we show the competitive performance of the proposed PRNN on four human action datasets. Our model provides an effective end-to-end solution for modeling temporal dynamics in action sequences by exploiting the PI in training time. Unlike most of the previous works that are based on a certain assumption about the structure of the depth maps or the availability of a robust skeleton tracker, our model automatically learns features from raw depth maps irrespective of any assumptions \cite{Yu_bmvc_2010,Wong_cvpr_2007} on the structure of video sequences .

\begin{table}[t]
\setlength{\tabcolsep}{0.4em}
\centering
\footnotesize
\scalebox{1.1}{\begin{tabular}{ l | l | l | l | l}
\hline



\multirow{1}{*}{Method} & Action3D & SBU & CAD60 & Blanket  \\ 
\hline
\hline
CNN-RNN (vanilla)  & 87.3 & 79.2  & 81.5 & 37.8 \\
\hline
PRNN-NoPreTrain  & 89.2 & 85.6 & 78.6 & 47.8 \\ 
\hline
PRNN-NoRefine  & 83.4  & 71.6  & 70.5 & 40.3   \\ 
\hline
\hline
PRNN  &   $\textbf{94.9}$  &   $\textbf{89.2}$  &   $\textbf{87.6}$ &  $\textbf{53.5}$ \\ 
\hline

\end{tabular}}
\caption{Contribution of each model component}
\label{tab:indiv}
\end{table}

\vspace{-0.1cm}
\subsection{Model Analysis}

\noindent \textbf{Evaluation of individual components} \ To verify the effect of individual components in our framework and demonstrate that if each of them contributes to the performance boost, we evaluate three variants of our approach: (1) PRNN-NoPreTrain discards the pre-training strategy as shown in Sec.~\ref{sec:pretrain}. Instead, the CNN encoder is trained from the scratch in the learning stage. (2) PRNN-NoRefine ignores the last refining step as described in Sec.~\ref{sec:refine}. The final model is trained by pre-train and learning steps. Note that the learning step in Sec.~\ref{sec:learning} can not be removed individually, because the latent PI is obtained based on the regression term of the learning step. We report the performance of a vanilla CNN-RNN pipeline. This is similar to our model in pre-training step, except that skeleton is not a part of the input during training. Note that our pre-training stage (taking both depth and skeleton as input) is specifically designed for our learning stage (with classification and regression loss). We tried to initialize vanilla CNN-RNN (depth input with classification loss) with our pretrained model. It performs much worse than learning from scratch. 



We show the average accuracy of all stripped-down versions of our model in Table~\ref{tab:indiv}. Overall, our method consistently achieves better performance with integrating each individual component, suggesting that each one of them contributes to the final performance. Without exploiting PI in the pre-training step, our model performs poorly due to the ineffective initialization. The vanilla CNN-RNN also suffers from the relatively small number of training data, and thus cannot take full advantage of the end-to-end manner. By considering the latent PI information in the refining step, this overfitting problem can be greatly alleviated from CNN-RNN and PRNN-NoRefine. It is clear that the performance has been substantially improved (PRNN) when combining these steps together. 




\begin{figure}[t]
\centering
   \includegraphics[width=\linewidth] {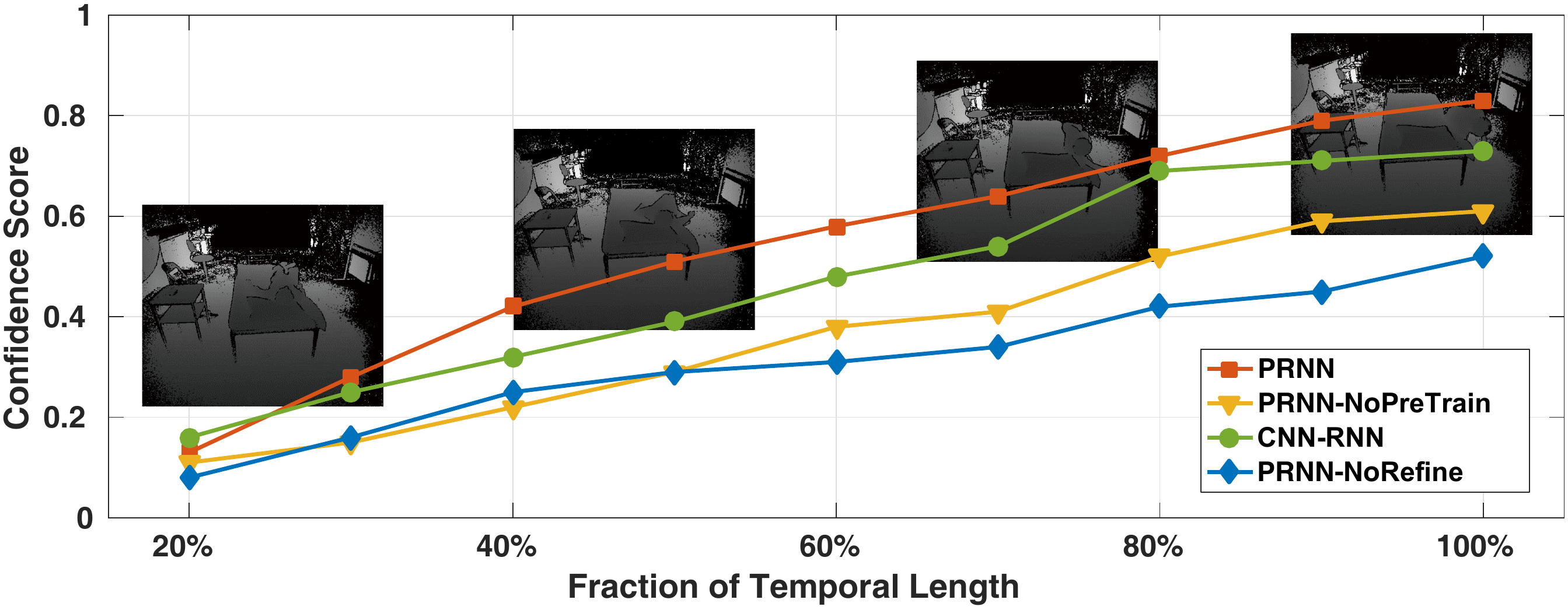}
 
\caption{Qualitative comparison of real-time prediction as time evolves for the action ``falling from the bed''}
\label{fig:ex_frame}
\end{figure}

\noindent \textbf{Qualitative analysis} We compare our approach with three variants in Figure~\ref{fig:ex_frame}, which illustrates the real-time prediction of an example sequence on every 15 time steps. The groundtruth action label is ``falling from the bed''. All methods give a low confidence to the correct action class at the beginning. As time evolves, we find that our approach first correctly predict the action labels. We attribute this faster learning ability to the mechanism of encoding PI \cite{Seungryul_wacv_2016}, which allows us to distinguish the subtle depth difference across successive frames.

\noindent \textbf{Computational efficiency} We take the Action3D as an example to discuss the efficiency of our approach.  With Python and C++ implementation on a NVIDIA Titan X GPU, our three-steps learning process takes about 11 hours to converge after continuously decreasing over 200k SGD iterations. Gradients are averaged over each minibatch in every training iteration. During testing, it can achieve real-time performance ($\approx$ 38 FPS). Compared with multi-stage models, the efficiency of our approach is mainly attributed to its end-to-end property without preprocessing step. Please refer to our supplementary video for real-time testing performance.

\vspace{-0.3cm}

\section{Conclusion and Future Work}
\vspace{-0.15cm}
\makeatletter{\renewcommand*{\@makefnmark}{}
\footnotetext{Acknowledgement: This work was supported by the Omron Corporation.}\makeatother} In this paper, we propose to learn a recurrent neural network with PI. The presented learning process provides threefold benefits: 1) The pre-training stage provides a mid-level embeddings which can be effectively tuned in the further stage. 2) In learning stage, a multi-task loss is formulated to exploit PI as a secondary task. 3) The learned information is further modeled to a latent PI, which is defined to close the gap between two tasks. The latent PI is used to enhance the discriminative power of the learned representation by closing two distributions. The latent PI is also updated iteratively in an EM fashion. In addition, the randomly sampled classification loss operates as a regularizer to reduce the tendency for overfitting. We apply our model to the problem of action recognition from depth sequences, and achieve better performance on three publicly available datasets and our newly collected dataset. In the future, we will consider to investigate more different types of PI and seek to model this information in the intermediate level of neural network \cite{JMLR_gulchere16a}.


{\small
\bibliographystyle{ieee}
\bibliography{egbib}
}

\end{document}